\documentclass{llncs}
\usepackage{graphicx}
\usepackage{epstopdf}
\usepackage{subfigure}
\usepackage{multirow}
\usepackage{amsmath}
\usepackage{url}
\usepackage{tikz}
\usepackage{balance}
\usepackage{hyperref}
\usepackage{adjustbox,lipsum}
\usepackage[ruled,vlined]{algorithm2e}
\usepackage[english]{babel}
\usepackage[utf8x]{inputenc}
\usepackage[T1]{fontenc}
\usepackage{float}
\begin{document}
\newcommand*{\textred}{\textcolor{red}}
\newcommand*{\textblue}{\textcolor{blue}}
\newcommand*{\textgreen}{\textcolor{green}}
\title{A Generalized Motion Pattern and FCN based approach for retinal fluid detection and segmentation}
\author{Shivin Yadav,Karthik Gopinath,Jayanthi Sivaswamy}

 \institute{IIIT Hyderabad,Hyderabad,India}


\maketitle              

\begin{abstract}
SD-OCT is a non invasive cross sectional imaging modality useful for diagnosis of macular defects. Efficient detection and segmentation of the abnormalities seen as biomarkers in OCT can help in analyzing the progression of the disease and advising effective treatment for the associated disease.
In this work we proposes a fully automated Generalized Motion Pattern(GMP) based segmentation method using a cascade of fully convolutional networks for detection and segmentation of retinal fluids from SD-OCT scans. General methods for segmentation depend on domain knowledge based feature extraction , whereas we propose a method based on Generalized   Motion   Pattern   (GMP) \cite{deepak2012detection}  which  is  derived  by  inducing  motion  to  an  image
to suppress the background.  The  proposed  method  is parallelizable and handles inter-scanner variability efficiently. Our method achieves a mean Dice score of 0.61,0.70 and 0.73 during segmentation and a mean AUC of 0.85,0.84 and 0.87 during detection for the 3 types of fluids IRF,SRF and PDE respectively. 
\end{abstract}
\begin{keywords}Generalized Motion Pattern,fluid-associated abnormalities,  retina, OCT.
\end{keywords}
\section{Introduction}

The main cause of blindness in developed countries are age related macular degeneration(AMD)\cite{bressler2004age} ,retinal vein occlusion\cite{hayreh1994retinal}  and diabetic maculopathy\cite{browning2008optical} . Retinal fluid (SRF and IRF) and sub-retinal pigment epithelium (sub-RPE) fluid(PED) are signs of age related macular degeneration and cystoidal macular edema thus their presence can act as a biomarker for early diagnosis of AMD and is helpful in analyzing prognosis of the disease and advising a treatment for the same. 

Spectral Domain OCT is a rapidly developing imaging modality which is effective in detection and quantization of cysts and sub retinal fluid abnormalities\cite{quellec2010three}. However, manual detection and segmentation of retinal fluids and sub-RPE fluids are laborious and time consuming. The presence, location, and extent of sub retinal fluid acts as disease biomarkers,  thus their volumetric quantification is beneficial for disease analysis, patient-tailored treatment  and  treatment  progress  assessment. Hence, there is a need for automated methods which gives accurate detection and quantization of the abnormalities.

We Propose a fully automatic method based on Generalized Motion Pattern for segmentation and detection of the retinal fluids using a cascaded Fully Convolutional Network(FCN)\cite{long2015fully} to form a joint segmentation and detection pipeline. FCN is shown to perform well in segmentation task across various modalities, Patrick et al.\cite{christ2016automatic} used a cascade of FCN to segment out liver and associated abnormalities achieving state of the art results in the same. The Generalized Motion Pattern helps in enhancement of the abnormalities such as retinal fluids and aids in speckle noise reduction. In this paper, a scanner independent method is developed by creating an ensemble of GMP's from the OCT scan and using this ensemble to perform our segmentation and detection task. This work is based on our previous work using a similar concept\cite{gopinath2017segmentation}. The details about the work is explained in section \ref{method} of this paper.

\section{Method and Data }
\label{method}

\begin{figure}
\caption{Pipeline of the proposed method}
\label{fig:pipeline}
\includegraphics[width=\linewidth]{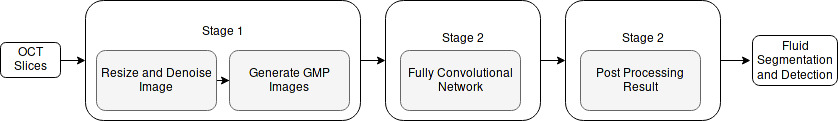}
\centering
\end{figure}

The pipeline for our method is shown in Fig.\ref{fig:pipeline} which is a comprises of 3 stages for segmentation and detection. In the first stage, we preprocess the data by denoising followed by resizing and ROI extraction. This data is used to generate the Generalized Motion Pattern images which forms the input to the cascaded FCN. The second stage in the pipeline is a cascade of fully convolutional networks for segmentation and detection of the retinal fluids. The final stage of the network involves post processing of the obtained prediction by refining the result. The details of the individual stages is explained in the subsections below .

\subsection{First Stage: Preprocessing and ROI extraction}
SD-OCT volumes are captured using different scanners and scanning protocols. Each scanner has varying intensity profile, and image resolution. In order to standardize data across scanners and reduce processing overhead we resize the image to 512$\times$256. The standardized volumes are then used to obtain a rough ROI region. We approach this problem by finding the brightest pixel in the slice along a column. Fitting a 1D Gaussian curve on the column wise projected vector for a volume, we obtaion the mean position of the brightest pixel. An ROI volume is extracted for the data by taking a margin around this pixel location. This reduced volume of size 256 $\times$256 reduces the overhead for the later stage.

OCT volumes contains varying speckle noise depending on the tissue captured. This noise often creates problems in functioning of various image processing algorithms. Traditional denoising algorithms  like median filtering and adaptive filtering based methods cannot preserve the boundary information. We use Spectral Total Variation based denoising \cite{chambolle2004algorithm} approach in this work because this method reduces the texture content and produces a smooth piecewise constant images preserving the edges. This denoised data is used as input for synthesizing GMP images.

 \subsubsection{Generalized Motion Pattern Images}
 \label{GMP}
The varying intensity and presence of different types of abnormalities across subjects and scanners makes automatic and accurate detection and segmentation task challenging. We propose a scheme to enhance the presence of an abnormality using Generalized Motion Pattern.
 
Given a gray scale image I, its GMP representation $I_{GMP}$ is defined as \begin{equation}
I_{GMP} (\vec{r}) = f(I(T_j (\vec{r})|1≤j≤N))
\end{equation}
Here $\vec{r}$ denotes the pixel location,$T_j(1≤j≤N)$ denotes $jth$ rigid  transformation  applied  to image I which  produces $jth$ resultant image. Total N such images are produced for each scan and these images are combined into the GMP map using a coalescing function $f(.)$, where $f(.)$ maps the set of pixel intensities at each location $\vec{r}$ across the transformed images to a scaler value.

For this challenge the rigid transformation chosen was translation. The translation is applied at an angle $\theta$ to the image in steps of $\delta$ from $-D$ to $D$ at different directions $\theta$. Hence, for translation in any direction $\theta$ we get a stack of $\frac{2D}{\delta}$ translated images along with the original image, forming a combined total of $\frac{2D}{\delta} +1$ images. The step size $\delta$ is set to be 1 and D to be 5 steps in this work. These images are then combined here using the coalescing function $minimum$ as the intensity profile for the retinal fluids is darker compared to its neighborhood.

Abnormalities appear in varying size and orientations. Translation along a single direction is insufficient for enhancing the abnormality region. Therefore, we propose to construct an ensemble of GMP images at various angles $\theta$ and enhance the presence of retinal fluids. In this paper we used $\theta$ as varying between $0^{\circ} and 180^{\circ}$ in steps of $22.5^{\circ}$ resulting in an ensemble of $K$ GMP images for each associated slice. This can be represented as\begin{equation}
C_k = \{I^k_{GMP}|1<k<K\}
\end{equation}
This ensemble of GMP images is combined by another coalescing function $\psi$ as \begin{equation}
I_{enhanced} = \psi(C_k)
\end{equation}
Volume correspondence helps in  extracting contextual information of the retinal fluids. The presence of fluid in one slice is a marker for the presence of similar fluid structures in neighboring locality across slices.
For introducing volume correspondence we propose using k neighboring slices in addition to the corresponding slice when constructing the GMP stack in our experiments we used k as 1 that is, using the slice preceding and the slice following the current slice .
 \subsection{Cascaded FCN Architecture}
 Using a predefined coalescing function ($\psi$) like $mean$, $max$ or $min$ is ineffective in enhancing only a particular type of abnormalities when the intensity of the surrounding is similar to the object of interest. Hence, there is a need to learn an optimal function $\psi$ capable of enhancing abnormalities of interests across subjects and scanners. 
 \begin{figure}[h]\includegraphics[width=\linewidth]{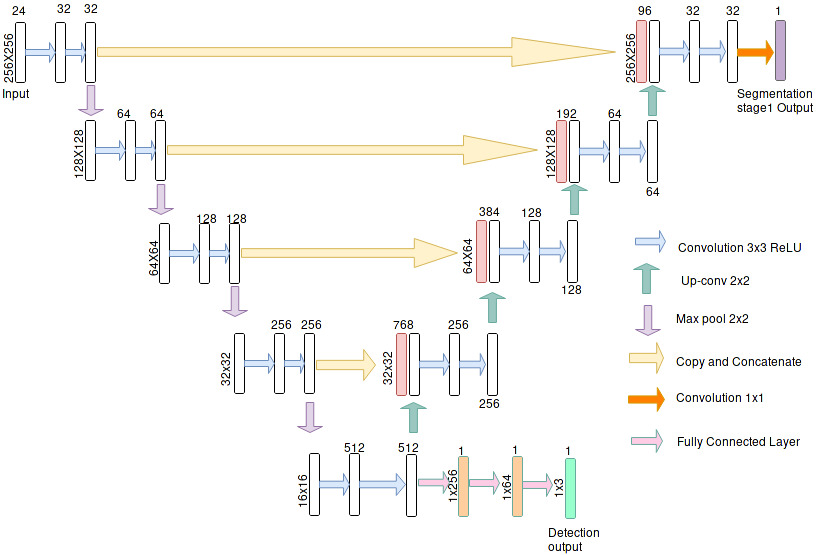}
\centering
\caption{FCN architecture used in both stages, for stage 1, we allow the network to learn the best function $\psi$ for combining the GMP ensemble to segment and detect the existence of retinal fluids, in stage 2 of the Cascaded FCN, the fluid prediction mask from stage1 and the original image are provided as inputs and the network produces a 3 different prediction masks for the 3 types of fluids}
\label{network}
\end{figure}
 
A CNN is generally used in computer vision tasks for solving classification and segmentation tasks. In this work we propose to use a CNN architecture to learn the function $\psi$ that will combine the ensemble of GMP images. The learned function $\psi$ will effectively map the ensemble of GMP images into an output image predicting the retinal fluids.
The design of the architecture for this problem is based on a cascade of Fully Convolutional Networks consisting of two independent networks joined in an end to end form to perform joint segmentation and detection. Both the independent networks here are similar to the widely used U-net FCN architecture\cite{DBLP:journals/corr/RonnebergerFB15}. The description of the network is shown in the Fig. \ref{network}.

The output of the first stage of the cascaded FCN is a map representing probable retinal fluid regions. The thresholded output of this map forms the retinal fluid prediction mask. GMP creates a smearing effect resulting in suppressing the edges of abnormalities with very less volume. To overcome this we provide the original image along with the predicted map as the input for the next stage of the cascaded network. The final FCN generates three masks for each type of retinal fluid as its output.
The detection subtask is handled by introducing a fully connected layers at the end of first cascaded stage. The predication at individual slice level for each type of retinal fluid is combined to produce a prediction for the entire volume as mentioned in section \ref{Post Processing}.

 \subsection{Post Processing}
\label{Post Processing}
The predicted regions from the FCN is sometimes plagued by presence of noise, creating false positives. The background region of some tissue structures resembling abnormalities are also enhanced by the GMP construction stage. These structures that affect the segmentation accuracy are removed during post processing.



 Thresholding the predicted map, we obtain segmented fluid regions as binary mask. Predicted regions having very few connected components are discarded as noise. Using this segmented mask and original image, we cluster the fluid regions in the intensity space removing the false positives.
Likewise for fluid detection, we threshold the slice wise prediction and detect abnormality in a slice by the gradient in probability measure. An increase or decrease in the slice wise probability of fluid indicates appearance or disappearance of abnormality across the volume. Since abnormalities are persistent 3D structures, considering k neighboring slices while predicting the presence of a fluid aids in accurate detection and helps eradicate false positives.

 \section{Experimental settings}
 \subsection{Dataset}
 The proposed method is evaluated on 70 SD-OCT volumes from 3 different OCT vendors Cirrus, Spectralis and Topcon. Each vendor data contains 3 sets with 8 volumes each. However, the third set from the Topcon vendor contains only 6 volumes.
 \subsection{Implementation details}
 The training and testing on the entire dataset is done using k-fold cross validation with k being eight. Our cascaded FCN was implemented using Keras library with Theano backend. Only the slices containing abnormalities were used for training the FCN with negative dice coefficient as the loss function. The first stage of the cascaded FCN was trained for 200 epochs and the second stage of the cascaded FCN was trained for 150 epochs for each fold on an Nvidia GTX-Titan X GPU.
 
 \subsection{Result Evaluation}
\begin{figure}[thbp]
 \centering
\begin{tabular}{ccc}
\centering
  \includegraphics[width=37mm]{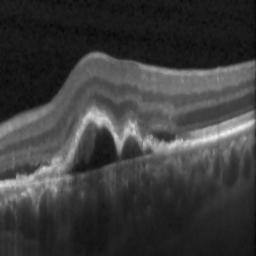} &
  \includegraphics[width=37mm]{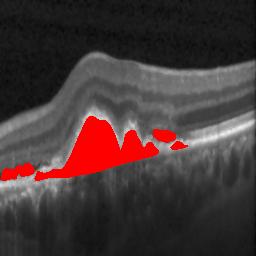}&
   \includegraphics[width=37mm]{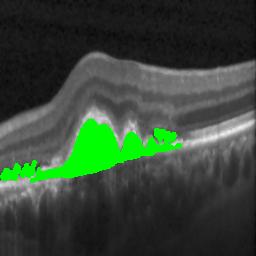}\\
(a) De-noised OCT slice & (b)Predicted fluid region & (c)Manual fluid segmentation \\[6pt]
  \includegraphics[width=35mm]{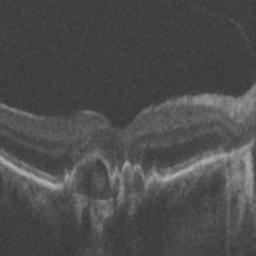} &
  \includegraphics[width=35mm]{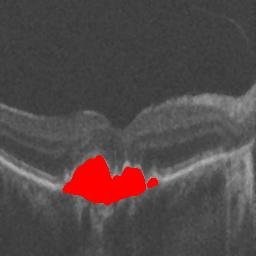}&
   \includegraphics[width=35mm]{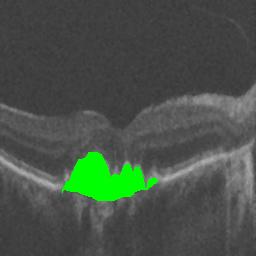}\\
(d) De-noised OCT slice & (e)Predicted fluid region & (f)Manual fluid segmentation \\[6pt]
 \includegraphics[width=35mm]{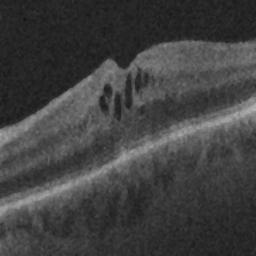} &
  \includegraphics[width=35mm]{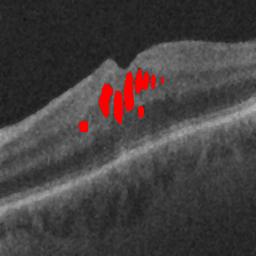}&
   \includegraphics[width=35mm]{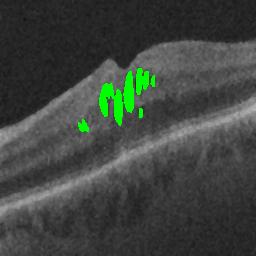}\\
(g) De-noised OCT slice & (h)Predicted fluid region & (i)Manual fluid segmentation \\[6pt]
    \includegraphics[width=35mm]{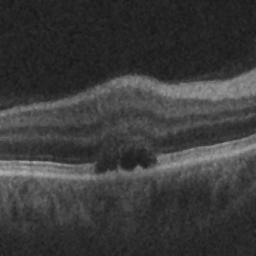} &
  \includegraphics[width=35mm]{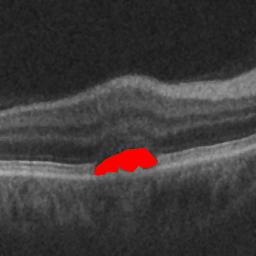}&
   \includegraphics[width=35mm]{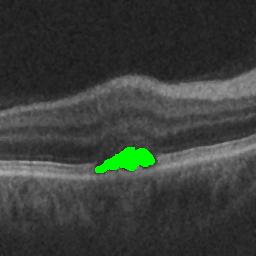}\\
(j) De-noised OCT slice & (k)Predicted fluid region & (l)Manual fluid segmentation \\[6pt]
    \includegraphics[width=35mm]{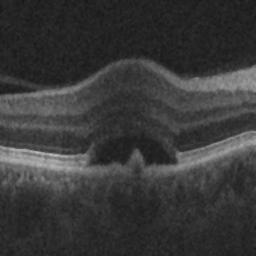} &
  \includegraphics[width=35mm]{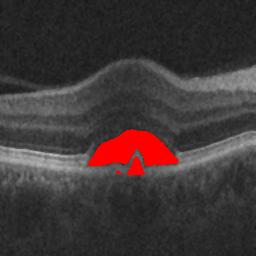}&
   \includegraphics[width=35mm]{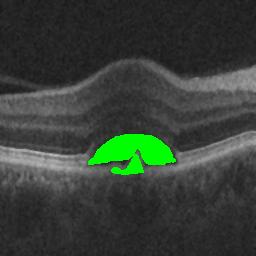}\\
(m) De-noised OCT slice & (n)Predicted fluid region & (o)Manual fluid segmentation \\[6pt]
\end{tabular}
\caption{Qualitative results for fluid segmentation  }
\label{fig:res}
\end{figure}

The qualitative results of the proposed system is shown in the Fig. \ref{fig:res}.
 The detection task is evaluated using Area Under the Curve(AUC) metric and the segmentation task is evaluated using Dice Coefficient(DC) metric and the results are presented in Table\ref{my-label}
 
\begin{table}[h]
\centering
\caption{Detection and Segmentation Results }
\label{my-label}
\resizebox{\linewidth}{!}{\begin{tabular}{|c|c|c|c|c|c|c|}
\hline
\multirow{2}{*}{Scanner Name} & \multicolumn{3}{c|}{AUC Score} & \multicolumn{3}{c|}{Dice Score} \\ \cline{2-7} 
 & IRF & SRF & PED & IRF & SRF & PED \\ \hline
Cirrus part1 & 0.67 & 0.90 & 0.92 & 0.66 & 0.82 & 0.69 \\ \hline
Cirrus part2 & 0.84 & 0.83 & 0.87 & 0.73 & 0.68 & 0.72 \\ \hline
Cirrus part3 & 0.82 & 0.84 & 0.87 & 0.61 & 0.75 & 0.79 \\ \hline
Spectralis part1 & 0.83 & 0.81 & 0.91 & 0.59 & 0.61 & 0.60 \\ \hline
Spectralis part2 & 0.82 & 0.88 & 0.89 & 0.47 & 0.85 & 0.82 \\ \hline
Spectralis part3 & 0.81 & 0.74 & 0.89 & 0.60 & 0.76 & 0.81 \\ \hline
Topcon part1 & 0.87 & 0.850 & 0.90 & 0.64 & 0.71 & 0.73 \\ \hline
Topcon part2 & 1 & 0.812 & 0.90 & 0.53 & 0.64 & 0.70 \\ \hline
Topcon part3 & 1 & 0.863 & 0.65 & 0.72 & 0.50 & 0.75 \\ \hline
Mean & 0.85 & 0.84 & 0.87 & 0.61 & 0.70 & 0.73 \\ \hline
\end{tabular}}
\end{table}

 


 \section{Conclusion}
 In this paper we presented a method to segment and detect retinal fluids in SD-OCT scans. Unlike segmentation methods which rely on domain based knowledge we presented a method which can be employed for segmenting a wide variety of abnormalities across different modalities. Inferring from the results, our method performed better on SRF and PDE compared to IRF in segmentation task due to the inherent nature of IRF resembling noise that gets enhanced while constructing the GMP. 
 
 The effects of this were not as profound in the detection stage as compared to the segmentation stage as the entire volume was taken into consideration when making prediction for the detection stage as compared to the segmentation stage which makes a prediction on a slice wise basis thus, small pockets of retinal fluids which are ignored as false positives do not affect the detection performance to the same extent as segmentation. 
 
 Methods based on domain knowledge such as location and intensity of layers can be taken into consideration in the post processing stage of the pipeline to adapt to these issues and a more advanced denoising algorithm based on local structures can also be adopted during the preprocessing stage to help improve the performance of the network.

\bibliographystyle{splncs}
\bibliography{Reference}

\end{document}